\documentclass[12pt,a4paper,twoside]{article}
\usepackage[utf8]{inputenc}
\usepackage[top=3cm,bottom=3cm,left=2.5cm,right=2.5cm]{geometry}
\usepackage[sf,bf]{titlesec} 

\title{{\LARGE\bf\sffamily Optimization Benchmark for Diffusion Models on Dynamical Systems\par}}
\author{Fabian Schaipp \\%
Inria, \'{E}cole Normale Sup\'{e}rieure, PSL Research University \\
Paris, France \\
\url{fabian.schaipp@inria.fr}
}

\date{}

\usepackage{xcolor}
\usepackage{amsmath,amssymb,amsfonts,amsthm}
\usepackage{mathtools}
\usepackage{bm}
\usepackage{dsfont}
\usepackage[]{algorithmic, algorithm}
\usepackage{enumitem}
\usepackage{natbib}
\usepackage{verbatim}

\usepackage{multirow}
\usepackage{longtable}
\usepackage{booktabs}

\usepackage[hidelinks,pagebackref]{hyperref}
\renewcommand*{\backref}[1]{}
\renewcommand*{\backrefalt}[4]{{\footnotesize [%
		\ifcase #1 Not cited.%
		\or Cited on page~#2%
		\else Cited on pages #2%
		\fi%
		]}}
	
\usepackage[capitalize,nameinlink]{cleveref}
\usepackage{subcaption} 
\usepackage{caption}

\usepackage{url}
\usepackage{cite}

\usepackage[parfill]{parskip}

\makeatletter
\def\thm@space@setup{%
	\thm@preskip=\parskip \thm@postskip=0pt
}
\makeatother

\let\temp\phi
\let\phi\varphi
\let\varphi\temp

\let\temp\varepsilon
\let\epsilon\varepsilon
\let\varepsilon\temp

\newcommand{\R}{\mathbb{R}}

\newcommand{\N}{\mathbb{N}}

\newcommand{\Adam}{{\texttt{Adam}}}
\newcommand{\SGD}{{\texttt{SGD}}}
\newcommand{\AdamW}{{\texttt{AdamW}}}
\newcommand{\SOAP}{{\texttt{SOAP}}}
\newcommand{\Shampoo}{{\texttt{Shampoo}}}
\newcommand{\Muon}{{\texttt{Muon}}}
\newcommand{\ScheduleFree}{{\texttt{ScheduleFree}}}
\newcommand{\Prodigy}{{\texttt{Prodigy}}}

\newcommand{\cosine}{\texttt{cosine}}
\newcommand{\wsd}{\texttt{wsd}}
\newcommand{\invsqrt}{\texttt{sqrt}}

\definecolor{kleinblue}{RGB}{0, 47, 167}
\definecolor{royalblue}{RGB}{0, 33, 115}
\hypersetup{
	colorlinks = True,
	linkcolor = black,
	citecolor=royalblue,
	urlcolor=royalblue
} 

\definecolor{todored}{RGB}{189, 30, 30}

\definecolor{nypink}{RGB}{216, 131, 115}
\definecolor{commblue}{RGB}{0, 66, 102}

\definecolor{takeawaycolor}{RGB}{155, 157, 137}
\definecolor{StrongRed}{RGB}{230, 57, 70}
\definecolor{OtherBlue}{RGB}{29, 112, 175}

\usepackage[colorinlistoftodos]{todonotes}

\newcommand{\numepochsavg}{five}
\newcommand{\emacoeff}{0.95}

\begin{document}
\maketitle


\begin{abstract}
  The training of diffusion models is often absent in the evaluation of new optimization techniques. In this work, we benchmark recent optimization algorithms for training a diffusion model for denoising flow trajectories. We observe that \Muon{} and \SOAP{} are highly efficient alternatives to \AdamW{} (18\% lower final loss). We also revisit several recent phenomena related to the training of models for text or image applications in the context of diffusion model training. This includes the impact of the learning-rate schedule on the training dynamics, and the performance gap between \Adam{} and \SGD{}.
\end{abstract}

\section{Introduction}

Over the last decade, the focus of optimization research has seen a shift towards applications in image classification and language modeling, particularly LLM pretraining. The training of \emph{diffusion models}, despite their impressive success and wide range of applications, is usually absent from empirical validation in optimization research. Even the most extensive efforts on optimization benchmarking \citep{Schmidt2021,Dahl2023,Kasimbeg2025a} do not contain results on diffusion models. Further, it remains unclear whether newly proposed methods, such as \SOAP{} \citep{Vyas2025} or \Muon{} \citep{Jordan2024}, are equally effective outside of LLM pretraining.

In this work, we validate whether recent trends in optimization for deep learning transfer to the training of diffusion models. In particular, our benchmark problem concerns training a diffusion model for denoising trajectories of dynamical systems, where the training data is obtained from fluid dynamics simulations. Our benchmark problem originally has been used for score-based data assimilation \citep{Rozet2023}; compared to the setting of LLM pretraining, it is different in terms of model architecture and loss function, data domain and training regime (multi/single epoch).

In order to run multiple seeds and hyperparameter configurations for all methods, our computational constraints only allow for relatively small-scale problems ($\sim$ 23M parameters).
Despite this limitation with respect to scale, the modeling technique from our benchmark problem has been successfully applied to diffusion-based data assimilation for regional and global weather and climate simulation \citep{Manshausen2024,Schmidt2025,Andry2025}. The findings of this benchmark might therefore be relevant and interesting to researchers who are training diffusion models for these scientific applications.

\paragraph{Benchmarking in optimization for machine learning.} The most extensive optimization benchmarking effort in recent years has been the \textit{AlgoPerf: Training Algorithms} benchmark \citep{Dahl2023, Kasimbeg2025a}. It consists of a variety of workloads, such as image classification and reconstruction, speech recognition, language translation, molecular property prediction, and click-through rate prediction. With \Shampoo{} \citep{Gupta2018,Anil2020} being one of the competition winners, the benchmark sparked renewed interest in dense matrix preconditioning techniques and led to the development of new algorithms, such as \SOAP{} \citep{Vyas2025} and \Muon{} \citep{Jordan2024}. \citet{Semenov2025} and \citet{Wen2025} recently conducted extensive benchmarking for LLM pretraining. Here, we study whether these new methods can also shine for our diffusion training task. In particular, we compare the performance of \SOAP{}, \Muon{} and \ScheduleFree{} \citep{Defazio2024} to the baseline method \AdamW{} \citep{Loshchilov2019}.

\paragraph{Performance gap between \Adam{} and \SGD{}.} In contrast to image classification with convolutional networks, where \SGD{} and \Adam{} perform equally well (if properly tuned), it is well known that \SGD{} does not easily%
\footnote{Recent works show that \SGD{} can close the gap to \Adam{} also for language tasks when using very small batch sizes \citep{Sreckovic2025,Marek2025}, or when applying \Adam{} only on the weights of the embedding layers \citep{Zhao2025}.
} achieve the same performance as \Adam{} for language modeling tasks \citep{Zhang2020b,Kunstner2023}.
\citet{Kunstner2024} further showed that imbalance of the class labels is sufficient to observe a gap between \Adam{} and \SGD{}. It remains unclear in which way other factors (for example, components of the model architecture) can have the same effect. Here, we investigate whether \SGD{} can close the gap to \Adam{} for an instance of diffusion model training, where the argument of class imbalance is not applicable.

\paragraph{Summary and main findings.} \Muon{} and \SOAP{} prove to be highly efficient also for diffusion model training. Despite their higher runtime per step compared to \AdamW{}, they achieve lower final loss values. \ScheduleFree{} almost matches \AdamW{} in terms of loss (without the need for scheduling), however we observe inferior generative quality. Similar effects can be observed for the \wsd{} schedule, which leads us to the conjecture that the entire training trajectory (and not only the final loss) is important for the quality of the trained diffusion model. 
We also observe a clear gap between \Adam{} and \SGD{}, which in this case can not be attributed to class imbalance.

\section{Experimental Setup}

Our experimental setup for training the diffusion model is following closely the setup of \citet{Rozet2023}: they train a \texttt{U-Net} model \citep{Ronneberger2015} which learns the score function of a dynamical system trajectory, obtained from the velocity field governed by the Navier-Stokes equations with Kolmogorov flow \citep{Kochkov2021}. Using the standard \texttt{DDPM} approach \citep{Ho2020}, the score function is learned by denoising data points sampled from the true distribution. We refer to \cref{sec:app:model-and-data} in the appendix for a detailed description of architecture and training data.

\paragraph{Hyperparameter tuning.} For each optimizer, we tune learning rate and weight decay separately (see \cref{fig:benchmark-heatmaps} for a detailed view on the grid search).
In general, we run three different seeds for each setting, and average all metrics across seeds.
If not specified otherwise, we run for $1024$ epochs with a linear-decay learning-rate schedule. Compared to \citet{Rozet2023}, we add warmup and gradient clipping by default (which lead to a minute reduction of the loss). A summary of the default hyperparameter settings is given in \cref{tab:default-hyperparams}. 


\paragraph{Computational cost.} A single run over $1024$ epochs with \AdamW{} takes roughly one hour one a single NVIDIA A100 GPU (this includes the end-of-epoch evaluations). In total, we executed $\sim$ 600 training runs, and utilized $\sim$ 830 A100-hours in total.
All experiments are conducted with \texttt{Pytorch} \citep{Paszke2019} of version $2.5.1$.
\section{Results}
\paragraph{Naming conventions.} We use \Adam{} and \AdamW{} interchangeably.
\ScheduleFree{} always refers to the \AdamW{} version presented by \citet{Defazio2024}.
%
%
\begin{figure}[t]
    \centering
    \begin{subfigure}{0.48\columnwidth}
    \includegraphics[width=0.99\textwidth]{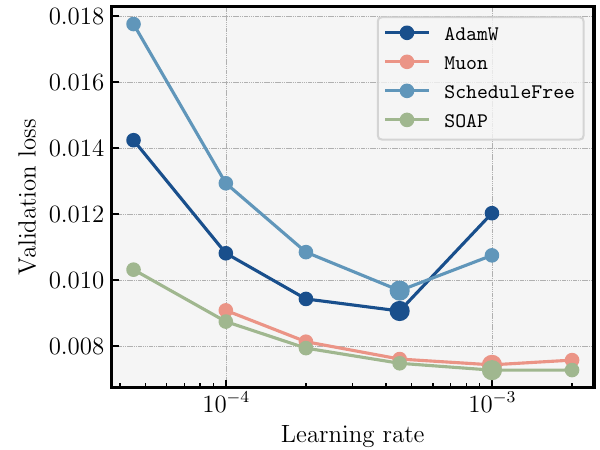}
    \end{subfigure}
    \begin{subfigure}{0.49\columnwidth}
        \includegraphics[width=0.99\textwidth]{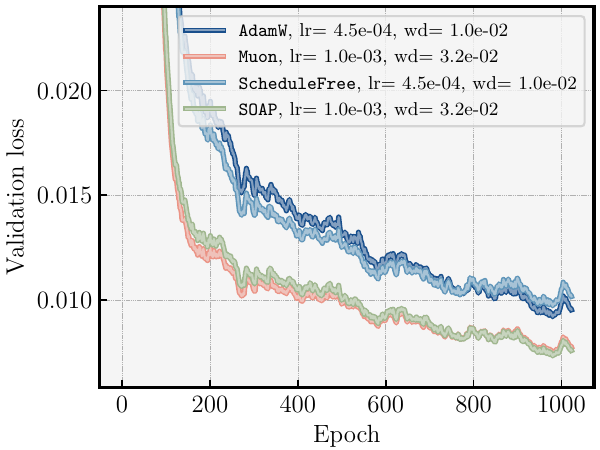}
    \end{subfigure}
    \caption{\textbf{(Left)} Final validation loss (averaged over the last \numepochsavg{} epochs) for each method and learning rate. Enlarged dot marks best learning rate. \textbf{(Right)} Validation loss curve for the best found setup for each method. Legend indicates learning rate (lr) and weight decay (wd) values. To obtain smoother curves we plot exponential moving averages with coefficient \emacoeff. See also \cref{fig:benchmark-2}.}
    \label{fig:benchmark}
\end{figure}
\subsection{Main Benchmark}
In this section, we compare the following optimizers:
\begin{itemize}[leftmargin=3ex]
    \item \AdamW{} \citep{Loshchilov2019}: Can be seen as the baseline method.
    \item \Muon{} \citep{Jordan2024}: Designed for $2$-dimensional weight matrices, and performs approximately steepest descent in the spectral norm \citep{Bernstein2025}. \Muon{} has been reported to improve convergence speed of LLM pretraining compared to \AdamW{} \citep{Liu2025}. See implementation details in \cref{sec:app:hyperparams}.
    \item \ScheduleFree{} \citep{Defazio2024}: An adaptation of \AdamW{} which does not require a learning-rate schedule (and therefore the length of training does not need to be pre-specified). We still use warmup, but afterwards the schedule is constant. \ScheduleFree{} won the self-tuning track of the \textit{AlgoPerf} benchmark \citep{Kasimbeg2025a}.
    \item \SOAP{} \citep{Vyas2025}: It combines the techniques from the \Shampoo{} algorithm and \Adam{}. \Shampoo{} won the external tuning track of the \textit{AlgoPerf} benchmark. As \SOAP{} is a subsequent development and has been reported to perform better, we opt to run \SOAP{} rather than \Shampoo{}.
\end{itemize}
\paragraph{Runtime per step.} It is important to point out that \Muon{} and \SOAP{} have a larger runtime per step than the other methods. In our setup, the training time of one epoch is roughly 1.45$\times$ larger for \Muon{} and 1.72$\times$ larger for \SOAP{} (compared to \AdamW{}). Given that runtime can significantly vary based on hardware and software setup, we focus on evaluation per steps, but also display loss curves with respect to training time.
We use publicly available implementations for \Muon{} and \SOAP{} and do not perform any software optimization in order to speed up these two methods specifically for our task.

\paragraph{Main results.} \cref{fig:benchmark} shows the final validations loss for each learning rate and method (here we pick the best weight decay setting for each point). The best performing run for each method is displayed on the right.
With respect to steps, \SOAP{} achieves the best performance, closely followed by \Muon{}. Over $1024$ epochs (equal to $26.6$K steps), \Muon{} and \SOAP{} achieve a loss value that is $18\%$ lower than the final loss of \AdamW{}.
\ScheduleFree{} improves over \AdamW{} early on in training, but falls slightly short in the end. With respect to runtime (see \cref{fig:benchmark-3}), \Muon{} performs best; \SOAP{} converges equally fast as \AdamW{}, but reaches a lower final loss. We stress that \textbf{the interpretation with respect to runtime might vary} based on hardware setup and software optimization.

\begin{figure}[t]
    \centering
    \begin{subfigure}{0.48\columnwidth}
    \includegraphics[width=0.99\textwidth]{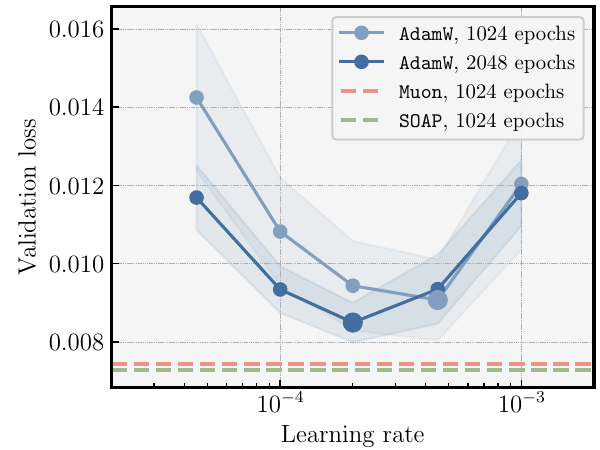}
    \end{subfigure}
    \begin{subfigure}{0.49\columnwidth}
        \includegraphics[width=0.99\textwidth]{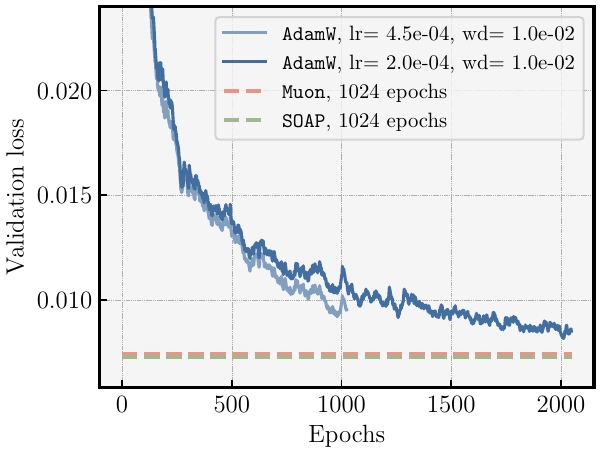}
    \end{subfigure}
    \caption{\textbf{(Left)} Final validation loss (averaged over the last \numepochsavg{} epochs, with band of one standard deviation over three seeds). Horizontal line marks best final loss for \Muon{} and \SOAP{} after 1024 epochs. \textbf{(Right)} Validation loss curve for the best \AdamW{} run over 1024 and 2048 epochs (smoothened by exponential moving averages with coefficient \emacoeff).}
    \label{fig:extended-adamw}
\end{figure}

\paragraph{What happens if we simply train \AdamW{} for longer?} When comparing in terms of runtime, the advantage of \Muon{} and \SOAP{} over \AdamW{} is reduced significantly. This leads to the question whether \AdamW{} can match the final loss of \SOAP{}/\Muon{} if we simply train for more epochs. \cref{fig:extended-adamw} shows that this is not the case. In this sense, \SOAP{} and \Muon{} achieve lower final loss values even with the same (or lower) runtime budget.
We should note that for the \AdamW{} runs over 2048 epochs, we only tune the learning rate, with weight decay fixed to $10^{-2}$. As sensitivity to weight decay is generally rather small, we do not expect this to impact the conclusion.

%

\paragraph{Mismatch of loss value and generative quality for \ScheduleFree{}.}
We find that specifically for \ScheduleFree{}, similar loss values do not correspond to similar quality of generated trajectories (see \cref{fig:benchmark-generated-images,fig:schedulefree-generated-images}). We conjecture that this is partially due to the missing learning-rate annealing: adding a linear cooldown to \ScheduleFree{} improves generative quality, at least for some hyperparameter configurations (\cref{fig:schedulefree-wsd-generated-images}).\footnote{Of course, adding a learning-rate scheduler defeats the original purpose of \ScheduleFree{}; we run this experiments rather to investigate whether the lack of cooldown is causing the issue.}

\begin{figure}[t]
    \centering
    \begin{subfigure}{0.48\columnwidth}
    \includegraphics[width=0.99\textwidth]{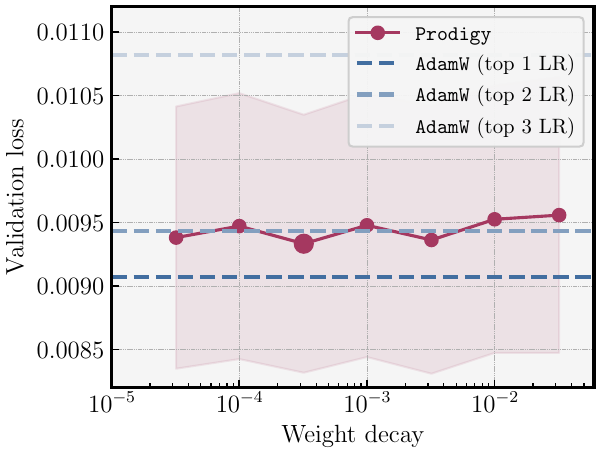}
    \end{subfigure}
    \begin{subfigure}{0.49\columnwidth}
        \includegraphics[width=0.99\textwidth]{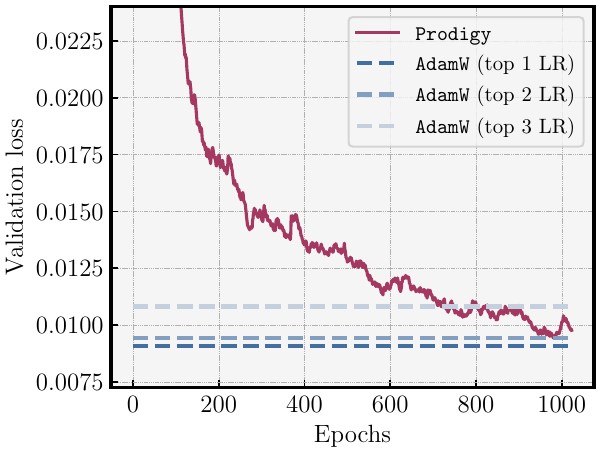}
    \end{subfigure}
    \caption{\textbf{(Left)} Final validation loss (averaged over the last \numepochsavg{} epochs) across weight decay for \Prodigy{}. Horizontal lines marks final loss for the three best learning-rates we have found for \AdamW{}. \textbf{(Right)} Validation loss curve for the best \Prodigy{} run (smoothened by exponential moving averages with coefficient \emacoeff).}
    \label{fig:prodigy}
\end{figure}

\paragraph{Can we avoid learning-rate tuning?}
We also try the \Prodigy{} optimizer proposed by \citet{Mishchenko2024}. They claim that \Prodigy{} automatically adapts to the optimal (peak) learning rate, and therefore only the schedule needs to be specified. We use the same linear-decay schedule as before, and tune weight decay with the same budget. \cref{fig:prodigy} shows that \Prodigy{} roughly matches the second-best learning-rate of \AdamW{} in terms of final loss, \emph{without any tuning} of the learning rate. 
The adaptive learning-rate of \Prodigy{} ramps up to a value of $6\cdot 10^{-4}$ within few epochs, which is reasonably close to the best learning rate we found for \AdamW{} through tuning. In terms of generative quality, the trajectories generated from the model trained with \Prodigy{} are visually of similar quality than the ones from \AdamW{} with tuned learning rate (see \cref{fig:benchmark-generated-images}).

\paragraph{Practical takeaways.} The optimal learning rate for \Muon{} and \SOAP{} is roughly twice as large as the optimal learning rate for \AdamW{}. We are confident that this is not problem-specific, as the same has been found by \citet{Semenov2025} for LLM training.
For our problem, sensitivity to the weight decay value is much smaller than to learning rate (see \cref{fig:benchmark-heatmaps}). Overall, \SOAP{} is the method that is least sensitive to learning rate/weight decay. Using the \Prodigy{} optimizer can reduce the tuning effort drastically with similar (or only to small extent inferior) model quality, which can be of practical advantage especially for preliminary training runs.

\subsection{Impact of Learning-Rate Schedule}
It is well-known that learning-rate schedules can drastically change training dynamics, in particular the shape of the loss curves; their behavior has been extensively studied for classical machine learning tasks on text or image data, see for example \citet{Defazio2023a,Haegele2024,Schaipp2025}. Here, we extend this to the training of diffusion models; we compare the effect of the schedule on the loss and the visual quality of the generated trajectories.\footnote{In the context of this section, \textit{schedule} refers to the learning-rate schedule, not the noise schedule of the diffusion model.}

\paragraph{Comparison of \wsd{} and \cosine{}.} A major drawback of the linear-decay (or \cosine{}) schedule is that the entire schedule depends on training length, which in consequence needs to be specified ahead-of-time. As an alternative, the \wsd{} schedule (``warmup-stable-decay'') has been proposed in the context of LLM pretraining: it keeps the learning rate constant, and a linear cooldown can be performed at any time  \citep{Zhai2022,Hu2024,Haegele2024}. The \wsd{} schedule matches or surpasses the performance of \cosine{} for LLM pretraining \citep{Haegele2024}.

\begin{figure}[t]
    \centering
    \begin{subfigure}{0.48\columnwidth}
        \includegraphics[width=0.99\textwidth]{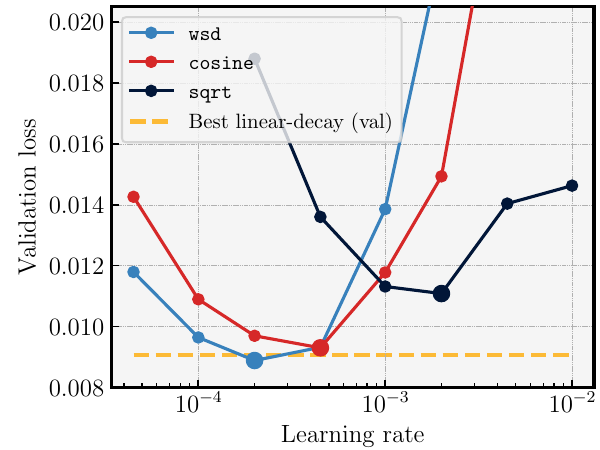}
        \caption*{\hspace{4ex}\textbf{(A)}}
    \end{subfigure}
    \begin{subfigure}{0.48\columnwidth}
        \includegraphics[width=0.99\textwidth]{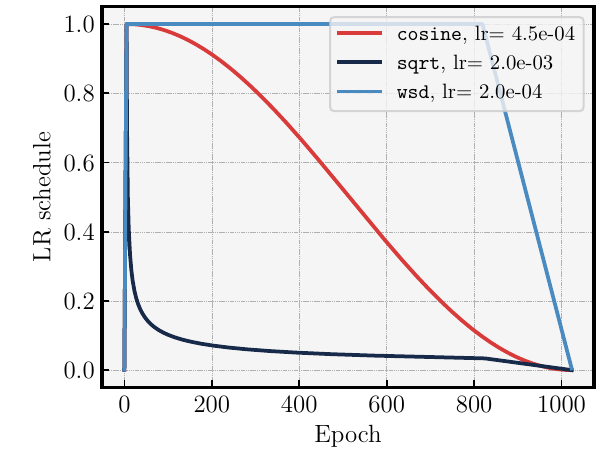}
        \caption*{\hspace{4ex}\textbf{(B)}}
    \end{subfigure}
    \begin{subfigure}{0.49\columnwidth}
        \includegraphics[width=0.99\textwidth]{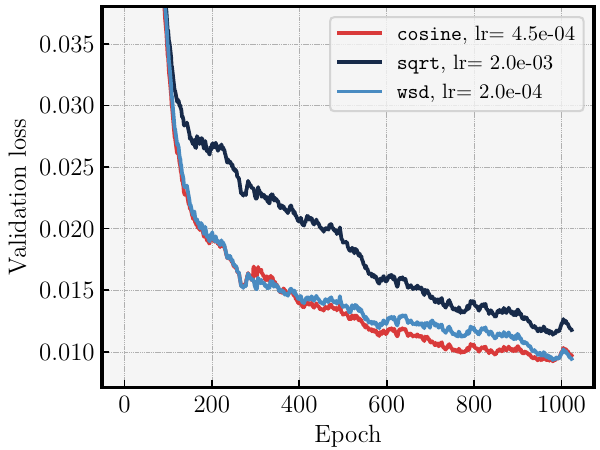}
        \caption*{\hspace{4ex}\textbf{(C)}}
    \end{subfigure}
    \begin{subfigure}{0.49\columnwidth}
        \includegraphics[width=0.99\textwidth]{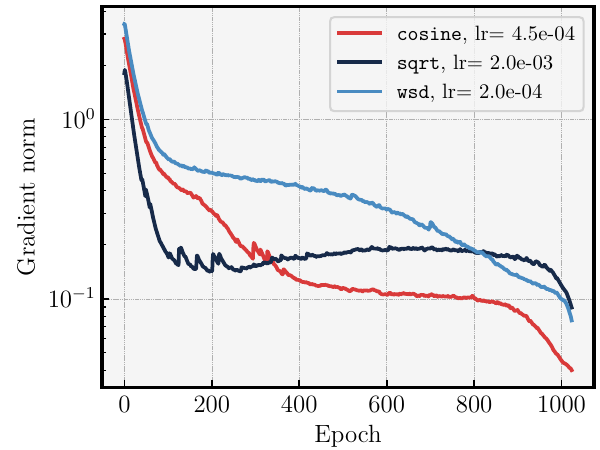}
        \caption*{\hspace{4ex}\textbf{(D)}}
    \end{subfigure}
    \caption{\textbf{(A)} Final validation loss (averaged over the last \numepochsavg{} epochs) for \cosine{}, \wsd{} and \invsqrt{} schedules across peak learning rate. Enlarged dot marks best learning rate. \textbf{(B)} Shape of the schedules, normalized by peak learning rate. For \wsd{} and \invsqrt{}, we use linear cooldown over the final 20\% of training. \textbf{(C)} Validation loss curve for the best found setup for each schedule (smoothened by exponential moving averages with coefficient \emacoeff).
    \textbf{(D)} Same as \textbf{(C)} for $\ell_2$-norm of the batch gradients.}
    \label{fig:schedule-comparison}
\end{figure}

Here, we find that, in terms of loss values, the same is true for the diffusion model training we consider (see \cref{fig:schedule-comparison}). Similar to empirical and theoretical findings by \citet{Haegele2024,Schaipp2025}, the optimal peak learning rate for \wsd{} is roughly half of the optimal one for \cosine{}. However, it seems that \textbf{generative quality becomes less stable when using the \wsd{} schedule} (see \cref{fig:schedules-generated-images}); for the learning rate that achieves minimal loss, the generated trajectories are of lower quality compared to the models trained with \cosine{} or linear-decay.

\paragraph{Alternative anytime schedule.} Motivated by the above shortcoming of the \wsd{} schedule, we try another ``anytime'' schedule, namely the inverse square-root schedule with linear cooldown \citep{Zhai2022}. As with \wsd{}, this schedule can be run -- except for the cooldown -- without specifying the length of training a priori. We refer from now on to this schedule with \invsqrt{}, a formal definition can be found in \cref{sec:app:hyperparams}.

\cref{fig:schedule-comparison} shows that the \invsqrt{} underperforms \wsd{} and \cosine{} in terms of final loss value. However, we observe that the quality of the generated trajectories is more stable than for \wsd{} (\cref{fig:schedules-generated-images}). Therefore, in situations where the training length cannot be specified ahead of time, the \invsqrt{} schedule appears to be a good alternative.

%
%

\subsection{Gap Between \AdamW{} and \SGD{}}

Here, we investigate whether there is a significant gap in training/validation loss between \AdamW{} and \SGD{}, when both methods are well-tuned. Starting from \citet{Kunstner2023}, this gap and its possible reasons have been studied extensively, mainly for image and language tasks. Our setup will add another perspective, as we study a different training task (diffusion), and data type (from turbulence simulation rather than images or text); in particular, the explanation that class imbalance causes the gap between \AdamW{} and \SGD{} can not be applied here, as there are no class labels involved.

\begin{figure}[t]
    \centering
    \begin{subfigure}{0.48\columnwidth}
    \includegraphics[width=0.99\textwidth]{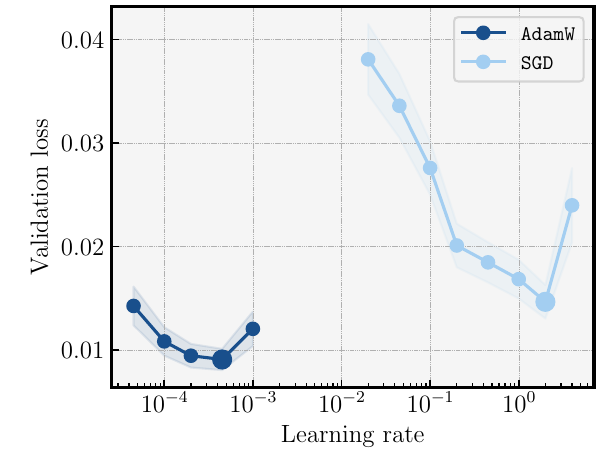}
    \end{subfigure}
    \begin{subfigure}{0.49\columnwidth}
        \includegraphics[width=0.99\textwidth]{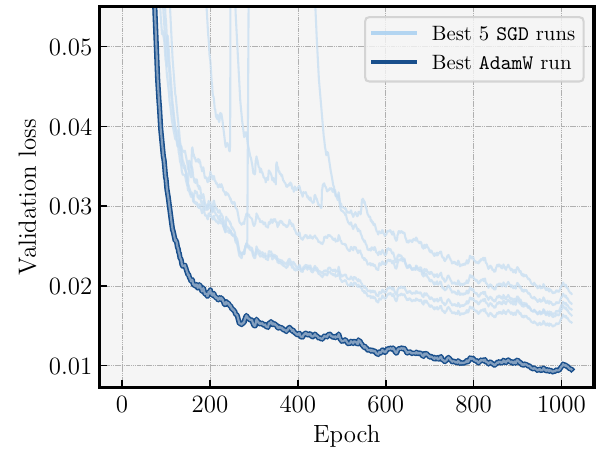}
    \end{subfigure}
    \caption{\textbf{(Left)} Final validation loss (averaged over the last \numepochsavg{} epochs) for each method and learning rate. \textbf{(Right)} Validation loss curve for the best found \AdamW{} setup, and the best five \SGD{} setups (smoothened by  exponential moving averages with coefficient \emacoeff).}
    \label{fig:adam-vs-sgd}
\end{figure}

\cref{fig:adam-vs-sgd} shows a significant gap in terms of validation loss between \AdamW{} and \SGD{}. For training loss, the results are qualitatively the same (plots not shown). The visual quality of the generated trajectories trained with \SGD{} are also clearly inferior, despite extensive hyperparameter tuning (see \cref{fig:sgd-additional-plots}).
This leads us to the conclusion that for this problem instance other factors (e.g.\ architecture properties) must be at play that explain the gap between \Adam{} and \SGD{}.

\section{Conclusion}
We show that \Muon{} and \SOAP{} are convincing alternatives to \AdamW{} for the training of diffusion models, even though their runtime per step is larger. Further, for our problem, we observed that for \ScheduleFree{} as well as for the \wsd{} learning-rate schedule the generative quality of the model can degrade, even though reasonably good loss values are achieved. We conjecture that the entire training trajectory might impact the final model quality, and leave this open for future work. 

Another open question that remains is to explain the performance difference between \Adam{} and \SGD{} and between \Muon{}/\SOAP{} and \Adam{}, as this benchmark problem lies outside of the domain of previously offered explanations.

\subsubsection*{Acknowledgments}
Fabian Schaipp is supported by the French government under the
management of Agence Nationale de la Recherche as part of the “Investissements d’avenir” program,
reference ANR-19-P3IA-0001 (PRAIRIE 3IA Institute), and the
European Research Council Starting Grant DYNASTY – 101039676.
This work was granted access to the HPC resources of IDRIS under the allocation 2025-AD011016024 made by GENCI.

\bibliography{lib}

\begin{thebibliography}{34}
\providecommand{\natexlab}[1]{#1}
\providecommand{\url}[1]{\texttt{#1}}
\expandafter\ifx\csname urlstyle\endcsname\relax
  \providecommand{\doi}[1]{doi: #1}\else
  \providecommand{\doi}{doi: \begingroup \urlstyle{rm}\Url}\fi

\bibitem[Andry et~al.(2025)Andry, Rozet, Lewin, Rochman, Mangeleer, Pirlet, Faulx, Grégoire, and Louppe]{Andry2025}
Gérôme Andry, François Rozet, Sacha Lewin, Omer Rochman, Victor Mangeleer, Matthias Pirlet, Elise Faulx, Marilaure Grégoire, and Gilles Louppe.
\newblock Appa: Bending weather dynamics with latent diffusion models for global data assimilation.
\newblock arXiv:2504.18720, 2025.

\bibitem[Anil et~al.(2020)Anil, Gupta, Koren, Regan, and Singer]{Anil2020}
Rohan Anil, Vineet Gupta, Tomer Koren, Kevin Regan, and Yoram Singer.
\newblock Scalable second order optimization for deep learning.
\newblock arXiv:2002.09018, 2020.

\bibitem[Bernstein \& Newhouse(2025)Bernstein and Newhouse]{Bernstein2025}
Jeremy Bernstein and Laker Newhouse.
\newblock Modular duality in deep learning.
\newblock In \emph{International Conference on Machine Learning}, 2025.

\bibitem[Carrassi et~al.(2018)Carrassi, Bocquet, Bertino, and Evensen]{Carrassi2018}
Alberto Carrassi, Marc Bocquet, Laurent Bertino, and Geir Evensen.
\newblock Data assimilation in the geosciences: An overview of methods, issues, and perspectives.
\newblock \emph{WIREs Climate Change}, 9\penalty0 (5):\penalty0 e535, 2018.

\bibitem[Dahl et~al.(2023)Dahl, Schneider, Nado, Agarwal, Sastry, Hennig, Medapati, Eschenhagen, Kasimbeg, Suo, Bae, Gilmer, Peirson, Khan, Anil, Rabbat, Krishnan, Snider, Amid, Chen, Maddison, Vasudev, Badura, Garg, and Mattson]{Dahl2023}
George~E. Dahl, Frank Schneider, Zachary Nado, Naman Agarwal, Chandramouli~Shama Sastry, Philipp Hennig, Sourabh Medapati, Runa Eschenhagen, Priya Kasimbeg, Daniel Suo, Juhan Bae, Justin Gilmer, Abel~L. Peirson, Bilal Khan, Rohan Anil, Mike Rabbat, Shankar Krishnan, Daniel Snider, Ehsan Amid, Kongtao Chen, Chris~J. Maddison, Rakshith Vasudev, Michal Badura, Ankush Garg, and Peter Mattson.
\newblock Benchmarking neural network training algorithms.
\newblock arXiv:2306.07179, 2023.

\bibitem[Defazio et~al.(2023)Defazio, Cutkosky, Mehta, and Mishchenko]{Defazio2023a}
Aaron Defazio, Ashok Cutkosky, Harsh Mehta, and Konstantin Mishchenko.
\newblock Optimal linear decay learning rate schedules and further refinements.
\newblock arXiv:2310.07831, 2023.

\bibitem[Defazio et~al.(2024)Defazio, Yang, Khaled, Mishchenko, Mehta, and Cutkosky]{Defazio2024}
Aaron Defazio, Xingyu Yang, Ahmed Khaled, Konstantin Mishchenko, Harsh Mehta, and Ashok Cutkosky.
\newblock The road less scheduled.
\newblock In \emph{Advances in Neural Information Processing Systems}, volume~37, pp.\  9974--10007, 2024.

\bibitem[Gupta et~al.(2018)Gupta, Koren, and Singer]{Gupta2018}
Vineet Gupta, Tomer Koren, and Yoram Singer.
\newblock Shampoo: Preconditioned stochastic tensor optimization.
\newblock In \emph{International Conference on Machine Learning}, volume~80, pp.\  1842--1850, 2018.

\bibitem[H\"{a}gele et~al.(2024)H\"{a}gele, Bakouch, Kosson, Ben~allal, Von~Werra, and Jaggi]{Haegele2024}
Alex H\"{a}gele, Elie Bakouch, Atli Kosson, Loubna Ben~allal, Leandro Von~Werra, and Martin Jaggi.
\newblock Scaling laws and compute-optimal training beyond fixed training durations.
\newblock In \emph{Advances in Neural Information Processing Systems}, volume~37, pp.\  76232--76264, 2024.

\bibitem[Ho et~al.(2020)Ho, Jain, and Abbeel]{Ho2020}
Jonathan Ho, Ajay Jain, and Pieter Abbeel.
\newblock Denoising diffusion probabilistic models.
\newblock In \emph{Advances in Neural Information Processing Systems}, volume~33, pp.\  6840--6851, 2020.

\bibitem[Hu et~al.(2024)Hu, Tu, Han, Cui, He, Zhao, Long, Zheng, Fang, Huang, Zhang, Thai, Wang, Yao, Zhao, Zhou, Cai, Zhai, Ding, Jia, Zeng, Li, Liu, and Sun]{Hu2024}
Shengding Hu, Yuge Tu, Xu~Han, Ganqu Cui, Chaoqun He, Weilin Zhao, Xiang Long, Zhi Zheng, Yewei Fang, Yuxiang Huang, Xinrong Zhang, Zhen~Leng Thai, Chongyi Wang, Yuan Yao, Chenyang Zhao, Jie Zhou, Jie Cai, Zhongwu Zhai, Ning Ding, Chao Jia, Guoyang Zeng, Dahai Li, Zhiyuan Liu, and Maosong Sun.
\newblock Mini{CPM}: Unveiling the potential of small language models with scalable training strategies.
\newblock In \emph{First Conference on Language Modeling}, 2024.

\bibitem[Jordan et~al.(2024)Jordan, Jin, Boza, Jiacheng, Cesista, Newhouse, and Bernstein]{Jordan2024}
Keller Jordan, Yuchen Jin, Vlado Boza, You Jiacheng, Franz Cesista, Laker Newhouse, and Jeremy Bernstein.
\newblock Muon: An optimizer for hidden layers in neural networks, 2024.
\newblock Blog post.

\bibitem[Kasimbeg et~al.(2025)Kasimbeg, Schneider, Eschenhagen, Bae, Sastry, Saroufim, Feng, Wright, Yang, Nado, Medapati, Hennig, Rabbat, and Dahl]{Kasimbeg2025a}
Priya Kasimbeg, Frank Schneider, Runa Eschenhagen, Juhan Bae, Chandramouli~Shama Sastry, Mark Saroufim, Boyuan Feng, Less Wright, Edward~Z. Yang, Zachary Nado, Sourabh Medapati, Philipp Hennig, Michael Rabbat, and George~E. Dahl.
\newblock Accelerating neural network training: An analysis of the algoperf competition.
\newblock In \emph{International Conference on Learning Representations}, 2025.

\bibitem[Kochkov et~al.(2021)Kochkov, Smith, Alieva, Wang, Brenner, and Hoyer]{Kochkov2021}
Dmitrii Kochkov, Jamie~A. Smith, Ayya Alieva, Qing Wang, Michael~P. Brenner, and Stephan Hoyer.
\newblock Machine learning–accelerated computational fluid dynamics.
\newblock \emph{Proceedings of the National Academy of Sciences}, 118\penalty0 (21):\penalty0 e2101784118, 2021.

\bibitem[Kunstner et~al.(2023)Kunstner, Chen, Lavington, and Schmidt]{Kunstner2023}
Frederik Kunstner, Jacques Chen, Jonathan~Wilder Lavington, and Mark Schmidt.
\newblock Noise is not the main factor behind the gap between {SGD} and {A}dam on transformers, but sign descent might be.
\newblock In \emph{International Conference on Learning Representations}, 2023.

\bibitem[Kunstner et~al.(2024)Kunstner, Milligan, Yadav, Schmidt, and Bietti]{Kunstner2024}
Frederik Kunstner, Alan Milligan, Robin Yadav, Mark Schmidt, and Alberto Bietti.
\newblock Heavy-tailed class imbalance and why {A}dam outperforms gradient descent on language models.
\newblock In \emph{Advances in Neural Information Processing Systems}, volume~37, pp.\  30106--30148, 2024.

\bibitem[Liu et~al.(2025)Liu, Su, Yao, Jiang, Lai, Du, Qin, Xu, Lu, Yan, Chen, Zheng, Liu, Liu, Yin, He, Zhu, Wang, Wang, Dong, Zhang, Kang, Zhang, Xu, Zhang, Wu, Zhou, and Yang]{Liu2025}
Jingyuan Liu, Jianlin Su, Xingcheng Yao, Zhejun Jiang, Guokun Lai, Yulun Du, Yidao Qin, Weixin Xu, Enzhe Lu, Junjie Yan, Yanru Chen, Huabin Zheng, Yibo Liu, Shaowei Liu, Bohong Yin, Weiran He, Han Zhu, Yuzhi Wang, Jianzhou Wang, Mengnan Dong, Zheng Zhang, Yongsheng Kang, Hao Zhang, Xinran Xu, Yutao Zhang, Yuxin Wu, Xinyu Zhou, and Zhilin Yang.
\newblock Muon is scalable for llm training.
\newblock arXiv:2502.16982, 2025.

\bibitem[Loshchilov \& Hutter(2019)Loshchilov and Hutter]{Loshchilov2019}
Ilya Loshchilov and Frank Hutter.
\newblock Decoupled weight decay regularization.
\newblock In \emph{International Conference on Learning Representations}, 2019.

\bibitem[Manshausen et~al.(2024)Manshausen, Cohen, Harrington, Pathak, Pritchard, Garg, Mardani, Kashinath, Byrne, and Brenowitz]{Manshausen2024}
Peter Manshausen, Yair Cohen, Peter Harrington, Jaideep Pathak, Mike Pritchard, Piyush Garg, Morteza Mardani, Karthik Kashinath, Simon Byrne, and Noah Brenowitz.
\newblock Generative data assimilation of sparse weather station observations at kilometer scales.
\newblock arXiv:2406.16947, 2024.

\bibitem[Marek et~al.(2025)Marek, Lotfi, Somasundaram, Wilson, and Goldblum]{Marek2025}
Martin Marek, Sanae Lotfi, Aditya Somasundaram, Andrew~Gordon Wilson, and Micah Goldblum.
\newblock Small batch size training for language models: When vanilla {SGD} works, and why gradient accumulation is wasteful.
\newblock arXiv:2507.07101, 2025.

\bibitem[Mishchenko \& Defazio(2024)Mishchenko and Defazio]{Mishchenko2024}
Konstantin Mishchenko and Aaron Defazio.
\newblock Prodigy: An expeditiously adaptive parameter-free learner.
\newblock In \emph{International Conference on Machine Learning}, volume 235, pp.\  35779--35804, 2024.

\bibitem[Paszke et~al.(2019)Paszke, Gross, Massa, Lerer, Bradbury, Chanan, Killeen, Lin, Gimelshein, Antiga, Desmaison, Kopf, Yang, DeVito, Raison, Tejani, Chilamkurthy, Steiner, Fang, Bai, and Chintala]{Paszke2019}
Adam Paszke, Sam Gross, Francisco Massa, Adam Lerer, James Bradbury, Gregory Chanan, Trevor Killeen, Zeming Lin, Natalia Gimelshein, Luca Antiga, Alban Desmaison, Andreas Kopf, Edward Yang, Zachary DeVito, Martin Raison, Alykhan Tejani, Sasank Chilamkurthy, Benoit Steiner, Lu~Fang, Junjie Bai, and Soumith Chintala.
\newblock Pytorch: An imperative style, high-performance deep learning library.
\newblock In \emph{Advances in Neural Information Processing Systems}, pp.\  8024--8035. 2019.

\bibitem[Ronneberger et~al.(2015)Ronneberger, Fischer, and Brox]{Ronneberger2015}
Olaf Ronneberger, Philipp Fischer, and Thomas Brox.
\newblock U-net: Convolutional networks for biomedical image segmentation.
\newblock In Nassir Navab, Joachim Hornegger, William~M. Wells, and Alejandro~F. Frangi (eds.), \emph{Medical Image Computing and Computer-Assisted Intervention -- MICCAI 2015}, pp.\  234--241, Cham, 2015. Springer International Publishing.
\newblock ISBN 978-3-319-24574-4.

\bibitem[Rozet \& Louppe(2023)Rozet and Louppe]{Rozet2023}
Fran\c{c}ois Rozet and Gilles Louppe.
\newblock Score-based data assimilation.
\newblock In \emph{Advances in Neural Information Processing Systems}, volume~36, pp.\  40521--40541, 2023.

\bibitem[Schaipp et~al.(2025)Schaipp, H\"{a}gele, Taylor, Simsekli, and Bach]{Schaipp2025}
Fabian Schaipp, Alexander H\"{a}gele, Adrien Taylor, Umut Simsekli, and Francis Bach.
\newblock The surprising agreement between convex optimization theory and learning-rate scheduling for large model training.
\newblock In \emph{International Conference on Machine Learning}, volume 267, pp.\  53267--53294, 2025.

\bibitem[Schmidt et~al.(2025)Schmidt, Schmidt, Strnad, Ludwig, and Hennig]{Schmidt2025}
Jonathan Schmidt, Luca Schmidt, Felix~M. Strnad, Nicole Ludwig, and Philipp Hennig.
\newblock A generative framework for probabilistic, spatiotemporally coherent downscaling of climate simulation.
\newblock \emph{npj Climate and Atmospheric Science}, 8\penalty0 (1), July 2025.

\bibitem[Schmidt et~al.(2021)Schmidt, Schneider, and Hennig]{Schmidt2021}
Robin~M Schmidt, Frank Schneider, and Philipp Hennig.
\newblock Descending through a crowded valley - benchmarking deep learning optimizers.
\newblock In \emph{International Conference on Machine Learning}, volume 139, pp.\  9367--9376, 2021.

\bibitem[Semenov et~al.(2025)Semenov, Pagliardini, and Jaggi]{Semenov2025}
Andrei Semenov, Matteo Pagliardini, and Martin Jaggi.
\newblock Benchmarking optimizers for large language model pretraining.
\newblock arXiv:2509.01440, 2025.

\bibitem[Srećković et~al.(2025)Srećković, Geiping, and Orvieto]{Sreckovic2025}
Teodora Srećković, Jonas Geiping, and Antonio Orvieto.
\newblock Is your batch size the problem? revisiting the {A}dam-{SGD} gap in language modeling.
\newblock arXiv:2506.12543, 2025.

\bibitem[Vyas et~al.(2025)Vyas, Morwani, Zhao, Shapira, Brandfonbrener, Janson, and Kakade]{Vyas2025}
Nikhil Vyas, Depen Morwani, Rosie Zhao, Itai Shapira, David Brandfonbrener, Lucas Janson, and Sham~M. Kakade.
\newblock {SOAP:} improving and stabilizing {S}hampoo using {A}dam for language modeling.
\newblock In \emph{International Conference on Learning Representations}, 2025.

\bibitem[Wen et~al.(2025)Wen, Hall, Ma, and Liang]{Wen2025}
Kaiyue Wen, David Hall, Tengyu Ma, and Percy Liang.
\newblock Fantastic pretraining optimizers and where to find them.
\newblock arXiv:2509.02046, 2025.

\bibitem[Zhai et~al.(2022)Zhai, Kolesnikov, Houlsby, and Beyer]{Zhai2022}
Xiaohua Zhai, Alexander Kolesnikov, Neil Houlsby, and Lucas Beyer.
\newblock Scaling vision transformers.
\newblock In \emph{Proceedings of the IEEE/CVF Conference on Computer Vision and Pattern Recognition (CVPR)}, pp.\  12104--12113, June 2022.

\bibitem[Zhang et~al.(2020)Zhang, Karimireddy, Veit, Kim, Reddi, Kumar, and Sra]{Zhang2020b}
Jingzhao Zhang, Sai~Praneeth Karimireddy, Andreas Veit, Seungyeon Kim, Sashank Reddi, Sanjiv Kumar, and Suvrit Sra.
\newblock Why are adaptive methods good for attention models?
\newblock In \emph{Advances in Neural Information Processing Systems}, volume~33, pp.\  15383--15393, 2020.

\bibitem[Zhao et~al.(2025)Zhao, Morwani, Brandfonbrener, Vyas, and Kakade]{Zhao2025}
Rosie Zhao, Depen Morwani, David Brandfonbrener, Nikhil Vyas, and Sham~M. Kakade.
\newblock Deconstructing what makes a good optimizer for autoregressive language models.
\newblock In \emph{International Conference on Learning Representations}, 2025.

\end{thebibliography}

\clearpage
\appendix

\tableofcontents

\section{Supplementary Material on Experiments}

We make the code and all training logs of this benchmark publicly available on~\href{https://github.com/fabian-sp/sda}{Github}. For our codebase, we have used the \href{https://github.com/francois-rozet/sda}{official implementation} of \citet{Rozet2023} as starting point.

\subsection{Overview of Model and Dataset}\label{sec:app:model-and-data}

\paragraph{Background on the learning task.}
Data assimilation is a central problem in many scientific domains that involve noisy measurements of complex dynamical systems, such as oceans or atmospheres (see \citet{Carrassi2018,Rozet2023} and references therein). Data assimilation can be seen as an inverse problem: the task is to estimate the distribution of true trajectories of the dynamical system, given a noisy measurement. The main contribution of \citet{Rozet2023} is to estimate this distribution based on a learned score function of true trajectories. This score function is obtained via standard diffusion model training. One advantage of their approach is that training and estimation can be performed entirely decoupled.
For our purpose of studying the performance of optimization algorithms, we focus solely on the training task.

\paragraph{Dataset.}
Our data generation procedure is identical to \citet{Rozet2023}. For the sake of completeness, we describe the main steps below. The input data for the diffusion model are snapshots of the (2-dimensional) velocity field which is governed by the Navier-Stokes equations. We follow \citet{Rozet2023,Kochkov2021} by solving the Navier-Stokes equations on a two-dimensional domain $[0,2\pi]^2$, with periodic boundary conditions, a large Reynolds number $Re = 1000$, a constant density, and an external forcing corresponding to Kolmogorov forcing with linear damping (cf.\ \citet{Kochkov2021}). The data is generated by solving 1024 independent trajectories of the Navier-Stokes equations (using \texttt{jax-cfd}) on a grid of resolution $256 \times 256$. Each trajectory consists of $128$ snapshots, which are then down-sampled to a resolution of $64 \times 64$, and filtered on the second half of the trajectory. We split the $1024$ trajectories into training (80\%), validation (10\%) and test (10\%) set.


During training, for each trajectory in the batch a random window of five snapshots is sampled with random starting point; this leaves us with input data of the shape $(b, 10, 64, 64)$, where $b$ is the batch size. 

\paragraph{Model architecture.} The model is a \texttt{U-Net} architecture \citep{Ronneberger2015} with three hidden convolutional layers, of channel sizes $(96, 192, 384)$. Further, we use a time embedding dimension of $64$. The model has $22.9$ million trainable parameters. For more details, we refer to \citet{Rozet2023}.

\subsection{Hyperparameters}\label{sec:app:hyperparams}

An overview of the default hyperparameters is given in \cref{tab:default-hyperparams}. Method-specific hyperparameter choices are listed thereafter.

\paragraph{Momentum coefficients.} For \SGD{}, we use heavy-ball momentum with coefficient $0.9$ (and \texttt{dampening} set to $0.9$). For \AdamW{}, \SOAP{}, and \ScheduleFree{}, we use always $(\beta_1, \beta_2) = (0.9, 0.999)$. For \Muon{}, see below.

\paragraph{Details on \Muon{} implementation.}
The core idea behind \Muon{} is, for a weight matrix with gradient $G \in \R^{d_1\times d_2}$, to compute (approximately) the closest orthogonal matrix $G$.  It is given by $UV^T$, where $G=U\Sigma V^T$ is the singular value decomposition \citep{Bernstein2025}. This poses the question how to trainable parameters that are not $2$-dimensional. Here, we follow the standard method proposed by \citet{Jordan2024}: all bias and (time) embedding parameters are optimized with \AdamW{}; for all parameters with more than two dimensions, we reshape their gradient into matrix shape, apply the Newton-Schulz algorithm, and reshape back to the original shape.\footnote{
This means that a parameters of shape $(d_0,\ldots,d_m)$ will be reshaped into the shape $(d_0, \prod_{j=1}^m d_j)$.}
Moreover, in order to avoid separate tuning of the learning rate and weight decay for the \AdamW{}-trained and the \Muon{}-trained parameters, we apply the heuristic of \citet{Liu2025}, which roughly aligns the update magnitude of the two methods, and therefore allows to use one single learning rate/weight decay.

For \Muon{}-trained parameters we use Nesterov momentum of $0.9$; for \AdamW{}-trained parameters we use $(\beta_1,\beta_2) = (0.9, 0.999)$.

\paragraph{Sampling hyperparameters.} We set all hyperparameters that are not directly related to the training algorithm exactly as \citet{Rozet2023}. In particular, they use a cosine schedule for the diffusion process. After training is completed, we sample two trajectories for 64 steps, always with the same seed.

\paragraph{Details on schedule comparison.}
We use epoch-wise schedulers, that is, the learning rate is unchanged over the course of each epoch.
If we decompose the learning rate into the schedule $(\eta_t)_{t\in \N}$ and a multiplicative factor $\gamma > 0$, then for each schedule $(\eta_t)_{t\in \N}$ we tune $\gamma$ independently. 
Without warmup, the formal definition of the schedules we consider is as follows: for $1\leq t \leq T+1$, let
\begin{align}\label{eqn:def-schedules}
    \eta_t^{\cosine{}} &= \frac12 (1+\cos(\frac{t-1}{T}\pi)) \\
    \eta_t^{\wsd{}} &= \begin{cases}
        1  \quad & 1 \leq t < T_0,\\
        1 - \frac{t-T_0}{T+1-T_0} \quad &T_0 \leq t \leq T+1,
    \end{cases} \\
    \eta_t^{\invsqrt{}} &= \begin{cases}
        \frac{1}{\sqrt{t}}  \quad & 1 \leq t < T_0,\\
        \frac{1}{\sqrt{T_0}} [1 - \frac{t-T_0}{T+1-T_0}] \quad &T_0 \leq t \leq T+1.
    \end{cases}
\end{align}

We add warmup by shifting the schedules given above to the right by 5 epochs (the length of warmup). For \wsd{} and \invsqrt{} we set $T_0 = \lfloor 0.8 T \rfloor = 819$, that is, the length of the cooldown amounts to 20\% of training. 

The tuning of $\gamma$ is displayed in \cref{fig:schedule-comparison}. For \wsd{} and \cosine{} schedules, we only tune $\gamma$ and keep weight decay fixed at $10^{-3}$ (the original setting in \citet{Rozet2023}). For \invsqrt{} we additionally try weight decay values $10^{-2}$ and $10^{-4}$.

\begin{table}[H]
    \centering
    \caption{Default hyperparameter settings (if not specified otherwise).}
    \label{tab:default-hyperparams}
    \begin{tabular}{ | c || c | c | } 
     \hline
     \textbf{Name} & \textbf{Default} & \textbf{Comment} \\
     \hline
     Warmup & $5$ epochs & not used in \citet{Rozet2023} \\ 
     Learning-rate schedule & linear-decay & \ScheduleFree{} uses warmup+constant. \\
     Gradient clipping & $1.0$ & not used in \citet{Rozet2023} \\
     Batch size & $32$ & - \\
     Epochs & $1024$ & - \\
     Momentum & $0.9$ & applies to \Muon{} and \SGD{} \\
     \AdamW{} Betas & $(0.9, 0.999)$ & applies to \AdamW{}, \Prodigy{}, \ScheduleFree{}, \SOAP{} \\
     \hline
    \end{tabular}
\end{table}
\begin{table}[H]
    \centering
    \caption{Method-specific hyperparameters for \Muon{}}
    \label{tab:muon-hyperparams}
    \begin{tabular}{ | c || c | } 
     \hline
     \textbf{Name} & \textbf{Value} \\
     \hline
     Nesterov momentum & true\\
     Newton-Schulz coefficients & $(3.4445, -4.7750,  2.0315)$ \\
     Newton-Schulz steps & 5 \\
     \hline
    \end{tabular}
\end{table}
\begin{table}[H]
    \centering
    \caption{Method-specific hyperparameters for \SOAP{}}
    \label{tab:soap-hyperparams}
    \begin{tabular}{ | c || c | } 
     \hline
     \textbf{Name} & \textbf{Value}\\
     \hline
     Preconditioning frequency & 10 \\
     Max preconditioning dimension & $10^5$ \\
     \hline
    \end{tabular}
\end{table}
\subsection{Additional Plots}\label{sec:app:additional-plots}

\begin{figure}[H]
    \centering
    \begin{subfigure}{0.49\columnwidth}
    \includegraphics[width=0.99\textwidth]{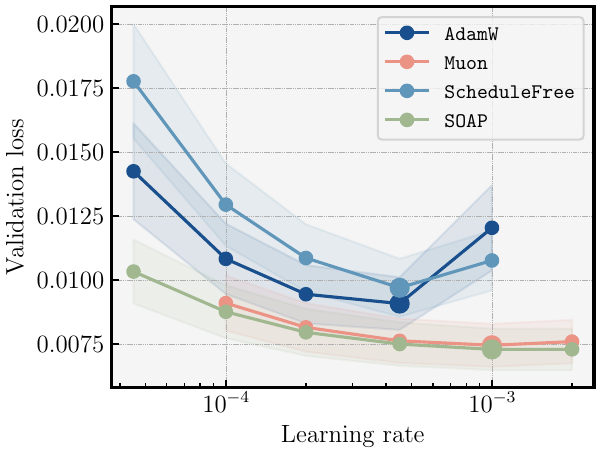}
    \end{subfigure}
    \begin{subfigure}{0.49\columnwidth}
    \includegraphics[width=0.99\textwidth]{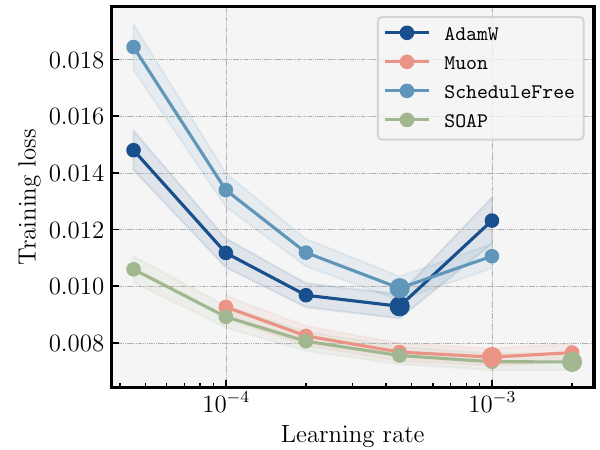}
    \end{subfigure}
    \caption{\textbf{(Left)} Same as \cref{fig:benchmark}, (left), but showing a band of one standard deviation over three runs. \textbf{(Right)} Same as (left), but for training loss.}
    \label{fig:benchmark-2}
\end{figure}
\begin{figure}[H]
    \centering
    \begin{subfigure}{0.49\columnwidth}
    \includegraphics[width=0.99\textwidth]{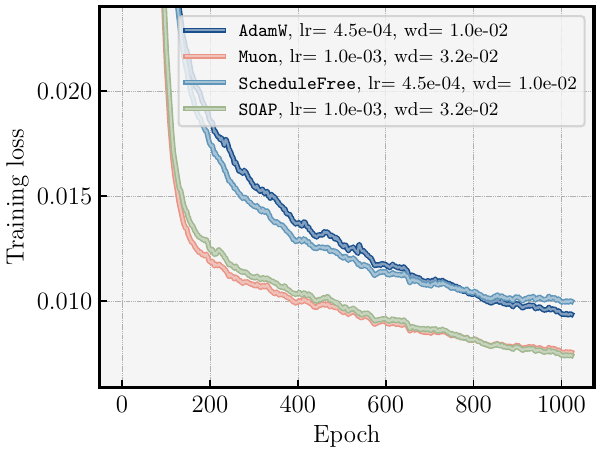}
    \end{subfigure}
    \begin{subfigure}{0.49\columnwidth}
        \includegraphics[width=0.99\textwidth]{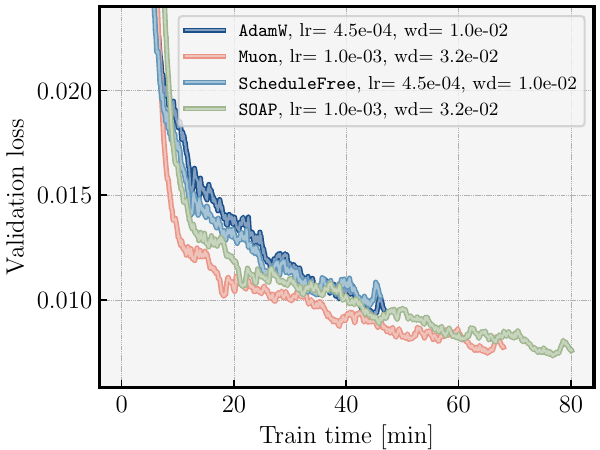}
    \end{subfigure}
    \caption{Training loss curve \textbf{(middle)} and validation loss curve with respect to train time for the best found setup for each method (minimal final validation loss). Legend indicates learning rate (lr) and weight decay (wd) values. To obtain smoother curves we plot exponential moving averages with coefficient \emacoeff.}
    \label{fig:benchmark-3}
\end{figure}
%

\begin{figure}[H]
    \centering
    \begin{subfigure}{0.49\columnwidth}
    \includegraphics[width=0.99\textwidth]{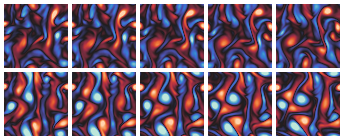}
    \caption{\AdamW{}}
    \end{subfigure}
    \begin{subfigure}{0.49\columnwidth}
    \includegraphics[width=0.99\textwidth]{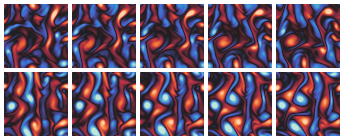}
    \caption{\Muon{}}
    \end{subfigure}
    \begin{subfigure}{0.49\columnwidth}
    \includegraphics[width=0.99\textwidth]{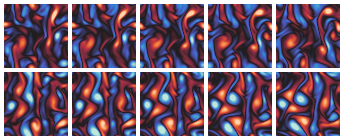}
    \caption{\SOAP{}}
    \end{subfigure}
    \begin{subfigure}{0.49\columnwidth}
    \includegraphics[width=0.99\textwidth]{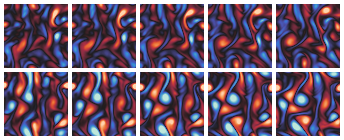}
    \caption{\Prodigy{}}
    \end{subfigure}
    \caption{Vorticity of the generated velocity field, plotted for two trajectories with five snapshots each, after training completed. For each method, we display the hyperparameters that achieved lowest validation loss.}
    \label{fig:benchmark-generated-images}
\end{figure}
%
\begin{figure}[H]
    \centering
    \begin{subfigure}{0.45\columnwidth}
    \includegraphics[width=0.99\textwidth]{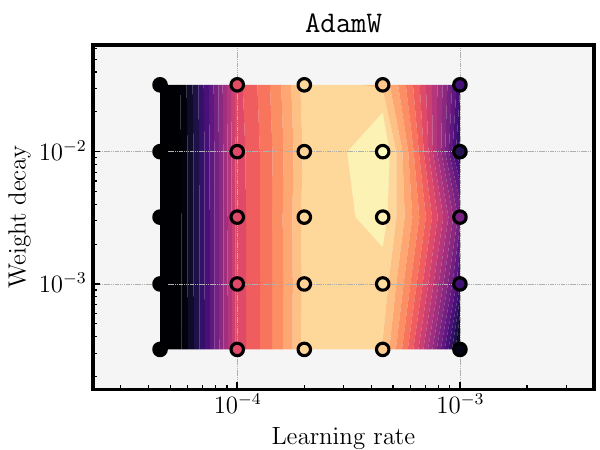}
    \end{subfigure}
    \begin{subfigure}{0.45\columnwidth}
        \includegraphics[width=0.99\textwidth]{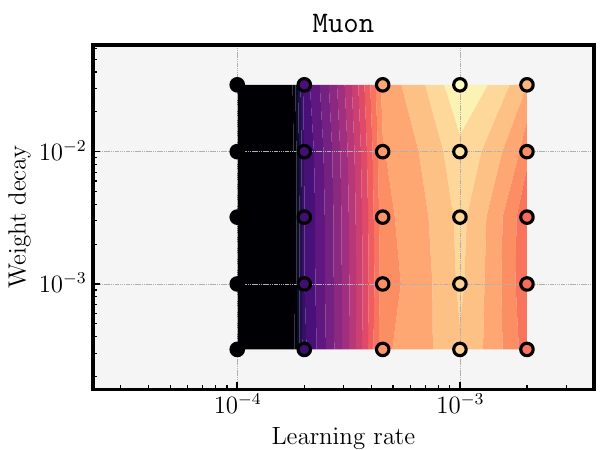}
    \end{subfigure}
    \begin{subfigure}{0.45\columnwidth}
    \includegraphics[width=0.99\textwidth]{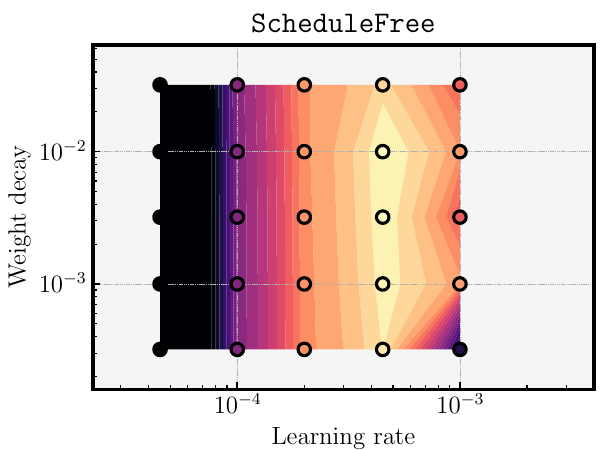}
    \end{subfigure}
    \begin{subfigure}{0.45\columnwidth}
        \includegraphics[width=0.99\textwidth]{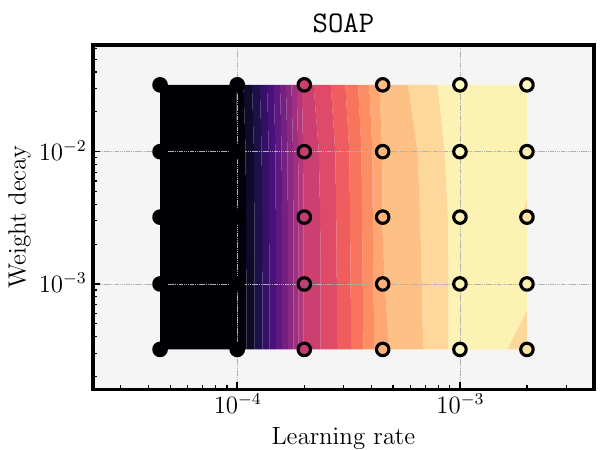}
    \end{subfigure}
    \caption{Heatmap of final validation loss (brighter is better) on the grid of learning rate and weight decay values. Each dot marks a hyperparameter combination that was run. Color indicates final validation loss (averaged over last \numepochsavg{} epochs), and color scale is different for each method in order to improve visibility.}
    \label{fig:benchmark-heatmaps}
\end{figure}
%
%
%
\begin{figure}[H]
    \centering
    %
    \begin{subfigure}{0.49\columnwidth}
    \includegraphics[width=0.99\textwidth]{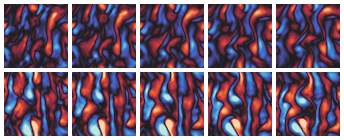}
    \caption{\wsd{}, lr=$2\cdot 10^{-4}$}
    \end{subfigure}
    \begin{subfigure}{0.49\columnwidth}
    \includegraphics[width=0.99\textwidth]{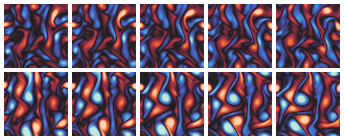}
    \caption{\wsd{}, lr=$4.5\cdot 10^{-4}$}
    \end{subfigure}
    %
     \begin{subfigure}{0.49\columnwidth}
    \includegraphics[width=0.99\textwidth]{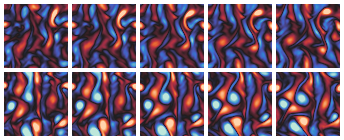}
    \caption{\invsqrt{}, lr=$1\cdot 10^{-3}$}
    \end{subfigure}
    \begin{subfigure}{0.49\columnwidth}
    \includegraphics[width=0.99\textwidth]{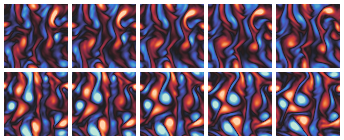}
    \caption{\invsqrt{}, lr=$2\cdot 10^{-3}$}
    \end{subfigure}
    %
    \begin{subfigure}{0.49\columnwidth}
    \includegraphics[width=0.99\textwidth]{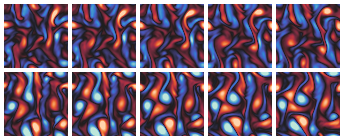}
    \caption{\cosine{}, lr=$2\cdot 10^{-4}$}
    \end{subfigure}
    \begin{subfigure}{0.49\columnwidth}
    \includegraphics[width=0.99\textwidth]{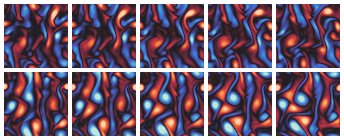}
    \caption{\cosine{}, lr=$4.5\cdot 10^{-4}$}
    \end{subfigure}
    \caption{Vorticity of the generated velocity field, plotted for the two best learning rates for each schedule. For \wsd{}, the learning rate that achieves minimal validation loss (a) actually results in lower quality of the generated trajectories. For \cosine{} and \invsqrt{} schedules this phenomenon does not occur. The finding is consistent across all three seeds.}
    \label{fig:schedules-generated-images}
\end{figure}
%

%
\begin{figure}[H]
    \centering
    \begin{subfigure}{0.49\columnwidth}
    \includegraphics[width=0.99\textwidth]{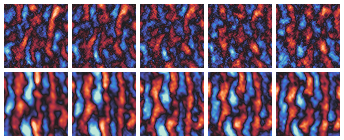}
    \caption{\ScheduleFree{} (train loss $0.00995$)}
    \end{subfigure}
    \begin{subfigure}{0.49\columnwidth}
    \includegraphics[width=0.99\textwidth]{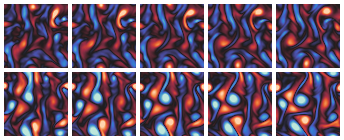}
    \caption{\AdamW{} (train loss $0.01022$)}
    \end{subfigure}
    \caption{\textbf{For \ScheduleFree{}, similar loss values do not result in similar generative quality.} Trajectories generated for the best \ScheduleFree{} run, and a \AdamW{} run with comparable, slightly higher, loss value. The quality of images generated with the model trained with \ScheduleFree{} is significantly worse.
    }
    \label{fig:schedulefree-generated-images}
\end{figure}
%
\begin{figure}[H]
    \centering
    \begin{subfigure}{0.49\columnwidth}
    \includegraphics[width=0.99\textwidth]{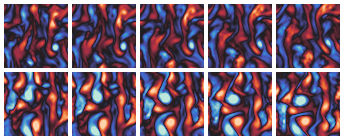}
    \caption{\ScheduleFree{} (train loss $0.01162$)}
    \end{subfigure}
    \begin{subfigure}{0.49\columnwidth}
    \includegraphics[width=0.99\textwidth]{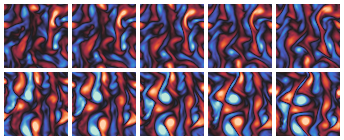}
    \caption{\ScheduleFree{} + \wsd{} (train loss $0.01136$)}
    \end{subfigure}
    \caption{\textbf{Learning-rate annealing on top of \ScheduleFree{} improves generative quality.}
    For \ScheduleFree{}, better loss values do not always correspond to better generative quality (compare \textbf{(left)} to \cref{fig:schedulefree-generated-images} (left)).
    \textbf{(Right)} When adding the \wsd{} schedule to \ScheduleFree{} with 20\% cooldown, the generative quality of the model improves (for some hyperparameter configurations). Here, we display learning rate=$0.001$ and weight decay=$0.00032$ (left and right).
    }
    \label{fig:schedulefree-wsd-generated-images}
\end{figure}
\begin{figure}[H]
    \centering
    \begin{subfigure}{0.52\columnwidth}
    \raisebox{10mm}{
    \includegraphics[width=0.99\textwidth]{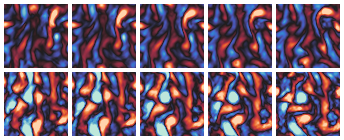}
    }
    \end{subfigure}
    \begin{subfigure}{0.44\columnwidth}
    \includegraphics[width=0.99\textwidth]{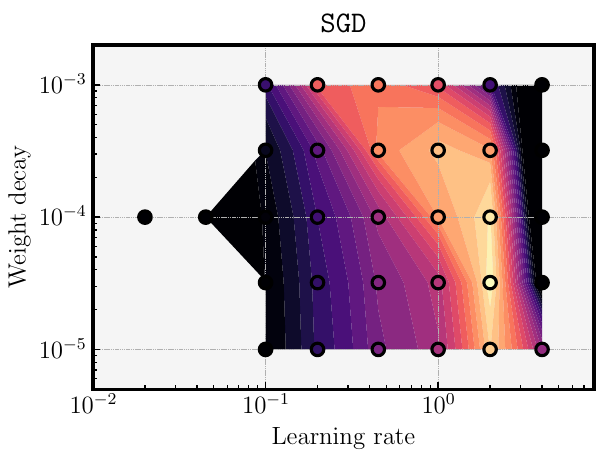}
    \end{subfigure}
    \caption{\textbf{(Left)} Vorticity of generated trajectories for the best setting we found for \SGD{}.  \textbf{(Right)} Heatmap of validation loss on the hyperparameter grid for \SGD{}, for details see caption of \cref{fig:benchmark-heatmaps}.
    }
    \label{fig:sgd-additional-plots}
\end{figure}

\end{document}